\documentclass{ecai} 

\usepackage{latexsym}
\usepackage{amssymb}
\usepackage{amsmath,mathtools}
\usepackage{amsthm}
\usepackage{dsfont}
\usepackage{booktabs,array}
\usepackage{enumitem}
\usepackage{graphicx}
\usepackage{color}
\usepackage{csquotes}
\usepackage{hyperref}
\usepackage{relsize}
\usepackage{forest} 

\newtheorem{definition}{Definition}
\newtheorem{assumptions}{Assumptions}
\theoremstyle{remark}
\newtheorem{remark}{Remark}

\newcommand{\BibTeX}{B\kern-.05em{\sc i\kern-.025em b}\kern-.08em\TeX}
\newcommand{\Concept}[1]{\ensuremath{\text{\normalfont\smaller\texttt{#1}}}}
\newcommand{\Fun}[1]{\ensuremath{\text{\normalfont\texttt{#1}}}}
\newcommand*{\result}[1]{\emph{#1}}

\DeclareMathOperator*{\argmin}{argmin}

\colorlet{leaf}{blue!80!black}
\tikzset{
    my node/.style={
      draw=gray,
      thin,
      rounded corners=3,
      text height=1.3ex,
      text depth=0ex,
      font=\sffamily,
      inner sep=2pt,
      text centered,
    },
    root node/.style={
        my node,
        draw=none,
        font=\itshape{root},
    },
    leaf node/.style={
      my node,
      draw=leaf,
      font=\sffamily\color{leaf},
    },
    strange node/.style={
      my node,
      inner color=red!5, outer color=red!80,
    },
  }
\forestset{
     default preamble={
     for tree={
        my node,
        l sep=0, s sep=1.5pt,
     },
 }
}

\begin{document}


\begin{frontmatter}


\paperid{123} 


\title{Unveiling Ontological Commitment in Multi-Modal Foundation Models}


\author[A,B]{\fnms{Mert}~\snm{Keser}\thanks{Corresponding Author. Email: mert.keser@continental.com}\footnote{Equal contribution.}}
\author[C]{\fnms{Gesina}~\snm{Schwalbe}
 \thanks{Corresponding Author. Email: gesina.schwalbe@uni-luebeck.de}\footnotemark}
\author[D]{\fnms{Niki}~\snm{Amini-Naieni}}
\author[E]{\fnms{Matthias}~\snm{Rottmann}}
\author[B]{\fnms{Alois}~\snm{Knoll}} 

\address[A]{Technical University of Munich, Germany}
\address[B]{Continental AG, Germany}
\address[C]{University of Lübeck, Germany}
\address[D]{University of Oxford, UK}
\address[E]{University of Wuppertal, Germany}




\begin{abstract}
Ontological commitment, i.e., used concepts, relations, and assumptions, are a corner stone of qualitative reasoning (QR) models.
The state-of-the-art for processing raw inputs, though, are deep neural networks (DNNs), nowadays often based off from multimodal foundation models. These automatically learn rich representations of concepts and respective reasoning. 
Unfortunately, the learned qualitative knowledge is opaque, preventing easy inspection, validation, or adaptation against available QR models.
So far, it is possible to associate pre-defined concepts with latent representations of DNNs, but extractable relations are mostly limited to semantic similarity. 
As a next step towards \emph{QR for validation and verification of DNNs}:
Concretely, we propose a method that \emph{extracts the learned superclass hierarchy} from a multimodal DNN for a given set of leaf concepts. 
Under the hood we 
(1) obtain leaf concept embeddings using the DNN's \emph{textual input modality};
(2) apply hierarchical clustering to them, using that \emph{DNNs encode semantic similarities via vector distances}; and
(3) label the such-obtained parent concepts using search in \emph{available ontologies from QR}.
An initial evaluation study shows that meaningful ontological class hierarchies can be extracted from state-of-the-art foundation models. Furthermore, we demonstrate how to validate and verify a DNN's learned representations against given ontologies.
Lastly, we discuss potential future applications in the context of QR.
\end{abstract}

\end{frontmatter}


\section{Introduction}
One of the basic ingredients of QR models is an ontology specifying the allowed concepts, relations, and any prior assumption about them; more precisely, the commitment to (a subset of an) ontology with associated semantic meaning of concepts and relations \cite{guarino1998formala}.
Thanks to years of research, large and rich ontologies like Cyc~\cite{lenat1989building}, SUMO~\cite{niles2001standard}, or ConceptNet~\cite{speer2017conceptnet}
are readily available for building or verifying QR models.

Meanwhile, however, DNNs have become the de-facto state of the art for many applications that hardly allow a precise input specification \cite{%
pouyanfar2018survey%
}, such as processing of raw images (\emph{computer vision}), e.g., for object detection \cite{%
grigorescu2020survey%
}, or processing of unstructured natural language text \cite{%
otter2021survey
}.
This machine learning approach owes its success to its strong representation learning capabilities:
DNNs automatically learn highly non-linear mappings (\emph{encoding}) from inputs to vectorial intermediate representations (\emph{latent representations} or vectors) \cite{%
caron2021emerging%
}, 
and reasoning-alike processing rules \cite{%
aghaeipoor2023fuzzy,%
he2020extract%
}
from these to a desired output.
Availability of large text and image datasets have further sparked the development of \emph{multimodal} so-called \emph{foundation models} \cite{%
bommasani2022opportunities,
kirillov2023segment,
radford2021learning
}.
These are large general-purpose DNNs trained to develop semantically rich encodings suitable for a variety of tasks \cite{bommasani2022opportunities}.
This is oft achieved by training them to map textual descriptions and images onto matching vectorial representations (\emph{text-to-image alignment}) \cite{%
radford2021learning
},
using multimodal inputs of both images and text.

\begin{figure}[t]
    \centering
    \vspace*{-1.5\baselineskip}%
    \includegraphics[width=.9\columnwidth]{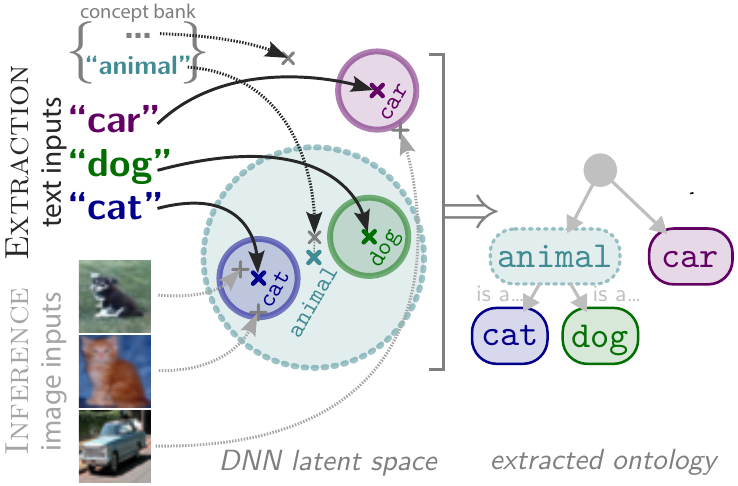}
    \caption{Illustration of the approach for ontology extraction from multimodal DNNs: For \emph{extraction}, (1) obtain leaf nodes (\Concept{cat}, \Concept{dog}, \Concept{car}) as the latent representations of their textual descriptions; (2) cluster these to get parent representations (\emph{dotted}); (3) assign parents the closest concept (\Concept{animal}) from a \emph{concept bank}. For \emph{inference} check at each level similarity against nodes' latent representations
    (e.g., first \Concept{animal} vs.\ \Concept{car}).}
    \label{fig:overview}
    \vspace*{1.5\baselineskip}
\end{figure}

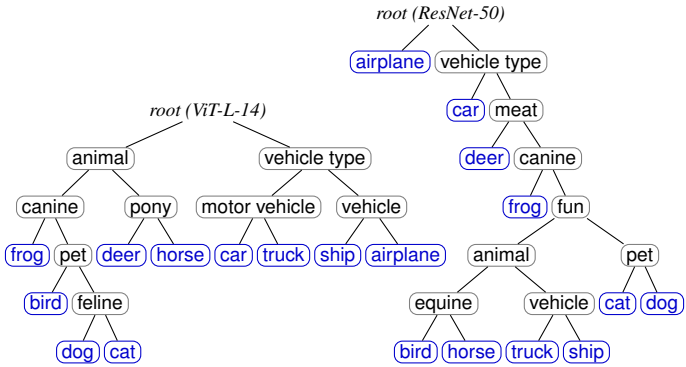
\begin{figure}[t]
\centering\scriptsize
\vspace*{-5\baselineskip}
\begin{forest}
    [~(ViT-L-14), root node
      [animal
        [canine
          [frog, leaf node]
          [pet
            [bird, leaf node]
            [feline
              [dog, leaf node]
              [cat, leaf node]
            ]
          ]
        ]
        [pony
          [deer, leaf node]
          [horse, leaf node]
        ]
      ]
      [vehicle type
        [motor vehicle
          [car, leaf node]
          [truck, leaf node]
        ]
        [vehicle
          [ship, leaf node]
          [airplane, leaf node]
        ]
      ]
    ]
\end{forest}%
\hspace*{-5em}%
\begin{forest}
    [~(ResNet-50), root node
      [airplane, leaf node]
      [vehicle type
        [car, leaf node]
        [meat
          [deer, leaf node]
          [canine
            [frog, leaf node]
            [fun
              [animal
                [equine
                  [bird, leaf node]
                  [horse, leaf node]
                ]
                [vehicle
                  [truck, leaf node]
                  [ship, leaf node]
                ]
              ]
              [pet
                [cat, leaf node]
                [dog, leaf node]
              ]
            ]
          ]
        ]
      ]
    ]
\end{forest}
\caption{
Comparison of two superclass hierarchies for given leaf concepts (\textcolor{leaf}{blue}) from CIFAR-10 \cite{alex2009learning} extracted from the large ViT-L-14 (\emph{left}; with optimized prompt; 92\% accuracy) and the smaller ResNet-50 (\emph{right}; 46\% accuracy) CLIP backbones with optimal distance metric settings.
It shows the positive influence of model quality and prompt optimization (using \textit{\enquote{a photo of a \Concept{class}}} instead of \enquote{\Concept{class}}) on the plausibility of the extracted ontology, and how the human-alignedness accuracy serves as indicator for it.
}
\label{fig:hierarchicalTree}
\vspace*{3\baselineskip}
\end{figure}

\paragraph{The prospect.}
Foundation models come with some interesting prospects regarding their learned knowledge:
(1) One can expect foundation models to \textbf{learn a possibly interesting and useful \emph{ontology}}, giving insights into \emph{concepts} \cite{%
kim2018interpretability,
lee2024neural,
schwalbe2022concept,
yuksekgonul2022posthoc
}
and concept relations \cite{%
fong2018net2vec,
kim2018interpretability
}
prevalent in the training data; and 
(2) such sufficiently large models can also \textbf{develop sophisticated reasoning chains} on the learned concepts \cite{%
he2020extract,
rabold2020expressive
}.
From the point of perspective of QR, this raises the question, whether this learned knowledge is consistent with the high quality available ontologies and QR models.
This opens up well-grounded verification and validation criteria for safety or ethically critical applications. As a first step towards this, this paper defines techniques for extraction and verification of simple class hierarchies.
Future prospects encompass to use the extracted knowledge from DNNs for knowledge retrieval, and
ultimately gain control over the learned reasoning: This would enable the creation of powerful \textbf{hybrid systems} \cite{%
donadello2017logic,
mao2018neurosymbolic
}
that unite learned encoding of raw inputs like images with QR models.

\paragraph{The problem.}
Unfortunately, the flexibility of DNNs in terms of knowledge representation comes at the cost of \emph{interpretability} \cite{%
gunning2019xai
};
and, being purely statistical models, they may extract \emph{unwanted and even unsafe correlations} \cite{%
kim2018interpretability,
ribeiro2016why,
schwalbe2020structuring
}.
The opaque distributed latent representations of the input do not readily reveal which interpretable concepts have been learned, nor what reasoning is applied to them for obtaining the output.
This is a pity, not least because that hinders verification of ethical and safety properties.
Take as an example the ontological commitment: Which hierarchical subclass-relations between concepts are considered? An example is shown in Fig.\,\ref{fig:example-commitment}.
This directly encodes the learned bias, which commonalities between classes are taken into account, and which of these are predominant for differentiating between classes.
The same example also nicely illustrates the issue with wrongly learned knowledge: The models may focus on irrelevant but correlated features to solve a task, such as typical background of an object in object detection \cite{ribeiro2016why}.
\begin{figure}[tbh]
{\scriptsize%
\begin{tabular}{l}
\toprule
(a) $\Concept{mammal}\supseteq\{\Concept{cat}, \Concept{dog}, \Concept{horse}\}$, 
    $\Concept{amphibian}\supseteq \{\Concept{frog}\}$%
\\\midrule
(b) $\Concept{indoor}\supseteq\{\Concept{cat}, \Concept{dog}\}, \Concept{outdoor}\supseteq\{\Concept{horse}, \Concept{wet} \}$, $\Concept{wet}\supseteq\{\Concept{frog}\}$
\\\bottomrule
\end{tabular}
}
\caption{Two exemplary ontological commitments: class hierarchies of the given leaf classes {\scriptsize\Concept{frog}, \Concept{cat}, \Concept{dog}, \Concept{horse}},
differentiating by
(a) biology (mammal vs.\ amphibian), 
(b) image background (a Clever Hans effect!).}
\label{fig:example-commitment}
\vspace*{1.5\baselineskip}
\end{figure}

A whole research field, \emph{explainable artificial intelligence} (XAI), has evolved that tries to overcome the lack of DNN interpretability \cite{%
gunning2019xai,
schwalbe2023comprehensive
}. 
To date it is possible to partly associate learned representations with interpretable symoblic \emph{concepts} (1-ary predicates) \cite{%
schwalbe2022enabling
}, 
such as whether an image region is a certain object part (e.g., \Concept{isLeg}), or of a certain texture (e.g., \Concept{isStriped}) \cite{%
fong2018net2vec,
kim2018interpretability
}.
However, extraction of learned relations is so far focused on simple semantic similarity of concepts \cite{%
fong2018net2vec,
schwalbe2021verification
};
hierarchical relations that hold across subsequent layers, i.e., across subsequent encoding steps \cite{%
kim2018interpretability,
wan2020nbdt,%
wang2020chain
};
or hierarchies obtained when \emph{sub}dividing a root concept \cite{mikriukov2023gcpv}.
And while first works recently pursued the idea to extract superclass hiearchies from given leaves, these are still limited to simple classifier architectures \cite{wan2020nbdt}.
A next step must therefore be: Given a set of (hierarchy leaf) concepts, how to extract (1) the \textbf{unifying superclasses}, and (2) the resulting \textbf{class hierarchy with subclass relationships} from any semantically rich intermediate output of a DNN, preferrably from the embedding space of \textbf{foundation models}.

\paragraph{Approach.}
We here propose a simple yet effective means to get hold of these encoded class hierarchies in foundation models;
thereby taking another step towards unveiling and verifying the ontological commitment of DNNs against known QR models respectively ontologies.
Building on \cite{wan2020nbdt} and \cite{yuksekgonul2022posthoc},
our approach leverages two intrinsic properties of the considered computer vision models:
\begin{enumerate}[label=(\arabic*)]
    \item Vision DNNs generally encode learned concept similarities via distances in their latent representation vector space \cite{fong2018net2vec}.
    This makes it reasonable to find a hierarchy of superclass representations by means of \textbf{hierarchical clustering} \cite{wan2020nbdt}.
    \item Foundation models accept textual descriptions as inputs, trained for \textbf{text-to-image alignment}. 
    This allows to cheaply establish an approximate bijection of textual concept descriptions to representations: A description is mapped by the DNN to a vector representation, and a given representation is assigned to that candidate textual description mapped to the most similar (=close by) vector \cite{yuksekgonul2022posthoc}.\footnote{%
    This could be replaced by the mentioned approximate concept extraction techniques for models without decoder and text-to-image alignment.}
\end{enumerate}

\paragraph{Contributions.}
Our main contributions and findings are:
\begin{itemize}[label=$\bigstar$]
\item An approach 
to \textbf{extract and complete a simple learned ontology}, namely a superclass hierarchy with given desired leaf concepts (\autoref{fig:hierarchicalTree}), from intermediate representations of any multimodal DNN, which allows to manually validate DNN-learned knowledge against QR models (see \autoref{fig:overview});
\item An approach to \textbf{test the consistency of multimodal DNNs against a given class hierarchy}, e.g., from standard ontologies;
\item An initial experimental validation showing that
    the approach can \textbf{extract meaningful ontologies},
    and reveal inconsistencies with given ontologies;
\item A thorough discussion of \textbf{potential applications} for QR extraction and insertion from / into DNNs.
\end{itemize}

\section{Related Work}

\paragraph{Extraction of learned ontologies.}
Within the field of XAI \cite{%
gunning2019xai,
schwalbe2023comprehensive
},
the subfield of concept-based XAI (c-XAI) has evolved around the goal to associate semantic concepts with vectors in the latent representations \cite{%
lee2024neural,
poeta2023conceptbased,
schwalbe2022concept
}.
For analysis purposes, methods here allow to both extract representations which match given concept specifications (supervised approach) \cite{%
fong2018net2vec,
keser2023interpretable,
kim2018interpretability,
yuksekgonul2022posthoc
}
as well as mine meanings for the most prevalent representations used by the DNN (unsupervised approach) \cite{%
ghorbani2019automatic,
zhang2021invertible
}.
Notably, we here utilize the supervised approach by Yuksekgonul et al.~\cite{yuksekgonul2022posthoc}
which directly utilizes the text-to-image alignment in multimodal DNNs.
Such associations have found manifold applications in the inspection of DNNs' learned ontology, such as:
Which concepts from a \emph{given} ontology are learned \cite{%
agafonov2022experiment,
schwalbe2022enabling
}?
And how similar are representations of different concepts \cite{%
fong2018net2vec,
schwalbe2021verification
}?
This was extended to questions about the QR of the models, such as 
sensitivity of later concept representations (or outputs) to ones in earlier layers \cite{%
kim2018interpretability
},
or compliance with pre-defined logical rules \cite{%
schwalbe2022enabling
}.
However, very few approaches so far explored more \emph{specific relations} between concept representations within \emph{the same} layer's representation space. In particular, specific relations beyond general semantic similarity, such as class hierarchies. This is a severe gap when trying to understand the learned ontological relations between concepts: 
DNNs develop increasing levels of abstraction across subsequent layers \cite{fong2018net2vec},
rendering the concepts occurring in their representation spaces hardly comparable.
Notably, Wan et al.~\cite{wan2020nbdt} challenged this gap and applied hierarchical clustering on DNN representations. However, their association of given concepts to latent representations is limited to last layer's output class representations, which we want to resolve. 
Furthermore, existing work was devoted only to single kinds of relations. We here want to show that these efforts can be unified under the perspective of investigating ontological commitment of DNNs.


\section{Background}

\subsection{Deep neural network representations}
\paragraph{DNNs.}
Mathematically speaking, deep neural networks are (almost everywhere) differentiable functions $F\colon\mathds{R}^n\to\mathds{R}^m$ 
which can be written in terms of small unit functions, the so-called \emph{neurons} $f\colon \mathds{R}^n\to\mathds{R}$, by means of 
the standard concatenation operation $f\circ g\colon x\mapsto f(g(x))$,
linear combination $x\mapsto Wx+b$,
and product $a, b\mapsto a\cdot b$.
Typically, the linear weights $W$ and biases $b$ serve as trainable parameters, which can be optimized in an iterative manner using, e.g., stochastic gradient descent.
Neurons are typically arranged in \emph{layers}, i.e., groups where no neuron receives outputs from the others. 
Due to this \enquote{Lego}-principle, DNNs are theoretically capable of approximating any continuous function (on a compact subspace) up to any desired accuracy \cite{%
hornik1991approximation
},
and layers can be processed highly parallel.
In practice, this is a double-edged sword: DNNs of manageable size 
show astonishing approximation capabilities for target functions like detection or pixel-wise segmentation of objects in images \cite{%
kirillov2023segment,
tan2020efficientdet
}.
However, they also tend to easily extract irrelevant correlations in the data, leading to incorrect \cite{ribeiro2016why}
or even non-robust \cite{szegedy2013intriguing}
generalization respectively \enquote{reasoning} on new inputs. 

\paragraph{Latent representations.}
In the course of an inference of an input $x$, each layer $L$ of the DNN produces as intermediate output a vector $F_{\to L}(x)\in\mathds{R}^{n}$, each entry being the output of one of the $n$ neurons of $L$.
This vectorial encoding of the input is called the \emph{latent representation} of the input within $L$, and the vector space $\mathds{R}^n$ hosting the representations is called the \emph{latent space}.
Interestingly, it was shown that DNNs encode semantically meaningful information about the input in their latent representations, with abstraction increasing the more layers are passed 
(e.g., starting with colors and textures, to later develop notions of shapes and objects) \cite{%
fong2018net2vec,
olah2017feature
}.

\paragraph{Concept embeddings.}
An emergent property of these representations is that in some layers, a concept $C$
(e.g., color \Concept{Red}, or object part \Concept{Leg}),
can be encoded as prototypical vector $e(C)$ within this latent space. These are called \emph{concept (activation) vectors} \cite{kim2018interpretability}
or \emph{concept embeddings} \cite{fong2018net2vec}.
The mapping $e\colon\mathcal{C}\to\mathds{R}^n$ from a set of human-interpretable concepts to their embeddings even preserves semantic similarities to some extend:
Examples are the reflection of analogical proportions \cite{prade2021analogical}
in word vector spaces (DNNs with textual inputs trained for natural language processing),
like
\enquote{$e(\Concept{King})-e(\Concept{Queen})=e(\Concept{Man})-e(\Concept{Woman})$}
\cite{%
mikolov2013linguistic
};
and their analogues in standard computer vision architectures trained for object classification or detection: \enquote{$e(\Concept{Green})+e(\Concept{Wood})=e(\Concept{Tree})$} \cite{fong2018net2vec}.
Our approach relies on these natural translation of semantic to vector operations/properties.
In particular, we assume that the relation $\Concept{IsSimilarTo}$\footnote{We here assume that \Concept{IsSimilarTo} is reflexive and symmetric, following geometrical instead of psychological models of similarity \cite{tversky1977features}.}
on input instances $x$ is mapped to some distance metric $d$ like Euclidean or cosine distance by the DNN representations:
$\forall \Concept{C}, \Concept{C'}\colon \Concept{IsSimilarTo}(\Concept{C}, \Concept{C'}) \Leftrightarrow d(e(\Concept{C}), e(\Concept{C'})\approx0$.\footnote{%
  For optimization, the relative formulation can be more convenient:
  \\$\forall \Concept{C},\Concept{C'}, \Concept{C''}\colon \Concept{C}\text{ more similar to }\Concept{C'}\text{ than to }\Concept{C''} \Rightarrow d(e(\Concept{C}),e(\Concept{\Concept{C'}}))\leq d(e(\Concept{C}),e(\Concept{\Concept{C''}}))$.}

Concretely, we use the translation of similarity relations to find a superclass concept representation via interpolation.

\paragraph{Text-to-image alignment.}
In the case of multimodal DNNs that accept both textual and image inputs, the training often encompasses an additional (soft) constraint:
Given textual descriptions of an input image, these must be mapped to the same/a similar latent representation as their respective image.
While pure language models suffer from the impossibility to learn the true meaning of language concepts without supervision \cite{%
bender2020climbing
},
this additional supervision might help the model to develop representations that better match the human understanding of the word/concept.
We here leverage this intrinsic mapping to associate textual or graphical descriptions of our concepts with latent representations. 

When using textual decriptions, good text-to-image alignment is an
important assumption; but, sadly, even with explicit training
constraints this is not guaranteed \cite{ge2023improving} (cf.\ distance of image and text embeddings in \autoref{fig:LatentClip}).
We show both the influence of text-to-image alignment on our method, how it can be reduced, and how to use our method in order to identify issues with the learned meaning of concepts, which opens up options to fix the representations.

\subsection{Ontologies}
When modeling any problem or world, a basis of the model is to know
\enquote{what the model is talking about}.
This is exactly answered by the underlying \emph{ontology}, i.e., a definition of what categories/properties and relations are used in the model.
We here adopt the definition from \cite{guarino1998formala}.
\begin{definition}[Ontology]\label{def:ontology}
  An \emph{ontology} is a pair $(\mathcal{V}, \mathcal{A})$ constituted by 
  a vocabulary $\mathcal{V}=\mathcal{C}\cup\mathcal{R}$ of 
  a set of unary predicates $\mathcal{C}$ (the \emph{concepts} corresponding to class memberships and other properties) and 
  a set of binary predicates $\mathcal{R}$ (the instance \emph{relations}) 
  used to describe a certain reality,
  and which are further constraint by a set $\mathcal{A}$ of explicit \emph{assumptions} in the form of a first- (or higher-)order logic theory on the predicates.
\end{definition}
A relation we will use further is $\Concept{IsSimilarTo}\in\mathcal{R}$. Also spatial relations like $\Concept{IsCloseBy}$ \cite{schwalbe2022enabling} and $\Concept{LeftOf}$, $\Concept{TopOf}$, etc. \cite{rabold2020expressive} have been defined and used in literature for latent space representations of objects. 
Simple examples of assumptions that relate the concept sets are, e.g., the subclass relationship we investigate in this paper: $\Concept{IsSuperclassOf}(\Concept{C'},\Concept{C}) :\Leftrightarrow (\forall v\colon \Concept{C}(v) \Rightarrow \Concept{C'}(v))$ (cf.\,\autoref{fig:example-commitment}).
This can also be seen as a relation between concepts, by interpreting the unary concept predicates $C$ as sets of objects (e.g., classes) via $v\in C :\Leftrightarrow C(v)$.
The validity of concept embeddings also gives rise to assumptions about concepts ($\forall v\colon C(v) \Leftrightarrow \Concept{IsSimilarTo}(v, e(C))$).
Note that, given embeddings, we can formulate relations between \emph{concepts} using \emph{instance} relations $R\in\mathcal{R}$ via $R(\Concept{C},\Concept{C'}) :\Leftrightarrow R(e(\Concept{C}), e(\Concept{C'}))$. An example would be $\Concept{isSimilarTo}(\Concept{cat},\Concept{dog})$.

The first challenge in extracting learned QR from DNNs is to find/explain the ontology that is used within the reasoning process of the DNN. Unraveling an ontology as done in \autoref{def:ontology} above breaks this step roughly down into:
\begin{enumerate}[label=(\arabic*),nosep, leftmargin=2em]
    \item Find the concepts $\mathcal{C}$ (and their embeddings) used by the model.
    \item Find the relations $\mathcal{R}$ that may be formulated on vector instances.
    \item Simple assumptions $\mathcal{A}_s\subseteq\mathcal{A}$: How are concept related.
    \item Identify further assumptions $\mathcal{A}\setminus\mathcal{A}_s$ that the model applies.
\end{enumerate}
Note that the layer-wise architecture of DNNs partitions the representations into objects (vectors) in the different latent spaces. For a layer $L$ we denote $\text{$v$ in the latent space of $L$}$ as $L(v)$.
This gives rise to a partition of the concept, relation, and assumption definitions, allowing to conveniently split up above steps as follows:
\begin{enumerate}[label=(\arabic*'), leftmargin=2em, nosep]
\item \label{step:layer-specific-concepts} What concepts $\mathcal{C}_i\subset\mathcal{C}$ are encoded \emph{within the $i$th layer} $L_i$
 \\($\forall C\in\mathcal{C}_i, v\colon \neg L_i(v)\Rightarrow \neg C(v)$)?
    \item[]\strut\vspace*{-\baselineskip}
        \begin{enumerate}[label=(3\alph*'), leftmargin=0em]
        \item\label{step:layer-specific-relations} What assumptions $\mathcal{A}_{i,i}$ hold for which items within \emph{the same} $i$th latent space
        ($\forall A\in\mathcal{A}_i,(v^{(s)})_{s}\colon \bigvee_s \neg L_i(v^{(s)}) \Rightarrow \neg A(v^{(1)}, \dots)$)?
        \item\label{step:cross-layer-relations} What assumptions $\mathcal{A}_{i,j}, i \neq j,$ hold between items of \emph{different} latent spaces?
    \end{enumerate}
\end{enumerate}
Task \ref{step:layer-specific-concepts} is (somewhat) solved by methods from c-XAI, where both learned concepts \cite{%
fong2018net2vec,
kim2018interpretability,
yuksekgonul2022posthoc
}
as well as their distribution over different layer representation spaces \cite{%
mikriukov2023revealing
}
are investigated.
\ref{step:layer-specific-relations} and \ref{step:cross-layer-relations} show the yet-to-be-filled gaps: Investigated relations between items, item groups respectively concepts within the same arbitrary latent space (=\ref{step:layer-specific-relations}). These so far only concern general semantic similarity, and relations across latent spaces only sensitivity. That falls far behind the richness of natural language; in particular it misses out on concept and instance relations of the kind \enquote{\Concept{C} is similar to \Concept{C'} with respect to feature \Concept{F}} respectively \enquote{\Concept{C}, \Concept{C'} both are \Concept{F}}, and counterpart \enquote{\Concept{C} differs from \Concept{C'} with respect to feature \Concept{F}}\footnote{%
\enquote{\Concept{C}, \Concept{C'} both are \Concept{F}} ($\forall x\colon(\Concept{C}(x)\vee\Concept{C'}(x))\Rightarrow\Concept{F}(x)$) rewrites to $\Concept{IsSuperclassOf}(F,C)\wedge\Concept{IsSuperclassOf}(F,C')$; the \enquote{differs}-case to $\neg\Concept{IsSuperclassOf}(F,C)\wedge\Concept{IsSuperclassOf}(F,C')$.
}.
In other words, the relation \Concept{IsSuperclassOf} is missing, despite known to be learned \cite{wan2020nbdt}.
This inhibits the expressivity of extracted constraints such as obtained in \cite{rabold2020expressive},
as this directly relies on the richness of available vocabulary.
The method proposed in this paper thus sorts in as follows: \textbf{We extend the extraction of relations relevant to point \ref{step:layer-specific-relations} (relations amongst concepts within the same layer representation space) by allowing to extract the \Concept{IsSuperclassOf} relation between concepts.}

\subsection{Hierarchical clustering}\label{sec:background-clustering}
Hierarchical clustering \cite{ran2023comprehensive} aims to find for a given set $M$ a chain of partitions $\mathcal{M}_1\leq\mathcal{M}_2\leq\dots\leq\{M\}$ connected by inclusion\footnote{%
To be precise: $\mathcal{M}\leq\mathcal{M}' \Leftrightarrow \forall M\in\mathcal{M}\colon\exists M'\in\mathcal{M}'\colon M\subseteq M'$
}, i.e., assign each point in $M$ to a chain of nested clusters $M_{1,i_1}\subseteq M_{2,i_2}\dots \subseteq M$, as illustrated in \autoref{fig:overview}.
Such a hierarchy can be depicted using a dendrogram as in \autoref{fig:hierarchicalTree}.
There are two regimes for hierarchical clustering: Divisive breaks up clusters top-down, while agglomerative starts from the leaves $\mathcal{M}_1=\{\{p\}\mid p\in M\}$ and iteratively merges clusters bottom-up \cite{ran2023comprehensive}.
We here employ hierarchical clustering to find a hierarchy of subsets of latent representation vectors. Since we start with given leaf vectors, \textbf{this work uses standard agglomerative hierarchical clustering \cite{wardjr.1963hierarchical}}.\footnote{We here use the scikit-learn implementation at \tiny\url{https://scikit-learn.org/stable/modules/generated/sklearn.cluster.AgglomerativeClustering.html}}
This optimizes the partitions for small distance between the \emph{single points} within a cluster (\emph{affinity}) and a large distance between the \emph{sets of points} making up different clusters (\emph{linkage}),
typically at a complexity of $\mathcal{O}(|M|^3)$.

\section{Approach}

This section details our approach towards extracting a globally valid approximation of a DNN's learned concept hierarchy, given the hierarchy's desired leaf concepts. The goal is to allow manual validation or verification testing against existing ontologies from QR.
Recall that this both requires a guided exploration of the learned concepts (\emph{which parent classes did the model learn?}), as well as an exploration of the applicability of the superclass relation (\emph{which superclasses/features are shared or different amongst given concepts?}).
We will start in \autoref{sec:approach-inference} by detailing how to obtain the extracted class hierarchy (here simply referred to as \emph{ontology}). This is followed by an excursion on how to conduct a kind of instance-based inference using the global taxonomy (\autoref{sec:approach-inference}, which is then used in \autoref{sec:approach-validation} where we discuss techniques for validation and verification of DNN learned knowledge.

\subsection{Extracting an ontology}\label{sec:approach-extraction}

\paragraph{Overview}
The steps to extract our desired ontology are (explained in detail further below):
(1) obtain the \textbf{embeddings} $e(\Concept{C}_i)$,
(2) apply \textbf{hierarchical clustering} to obtain superclass representations as superclass cluster centers,
(3) \textbf{decode} the obtained superclass representations into a human-interpretable description.

\paragraph{Ingredients.}
We need as ingredients
our trained \textbf{DNN} $F$,
some concept encoder $e$ (in our case defined using the DNN, see Step 1 below),
the finite set $(\Concept{C}_i)_i=\mathcal{C}_{\text{leaf}}$ of \textbf{leaf concepts} for which we want to find parents classes, and
the choice of \textbf{layer} $L$ in which we search for them.
Furthermore, to ensure human interpretability of the results, we constrain both our leaf concepts as well as our solution parent concepts to come from a given \textbf{concept bank} $\mathcal{C}$ of human-interpretable concepts\footnote{%
The concept bank restriction makes this essentially a search problem.}.
We furthermore need per concept $\Concept{C}\in\mathcal{C}$:
A \textbf{textual description} $\Fun{toText}(\Concept{C})$ of $\Concept{C}$ as textual specification;
optionally a set $\Fun{toImages}(\Concept{C})$ containing the concept as graphical specification (see Step 1), as available, e.g., from many densely labeled image datasets \cite{%
bau2017network,
he2022partimagenet
}; and
optionally a set $\Fun{Parents}(\Concept{C})$ of candidates for parent concepts of $\Concept{C}$ (for more efficient search).
The following assumptions must be fulfilled, in order to make our approach applicable:
\begin{assumptions}\label{assumptions}~
  \begin{enumerate}[nosep, label=(\alph*)]
  \item \textbf{Text-to-image alignment:}\label{ass:text-to-image-alignment}
    The DNN should accept textual inputs, and be trained for text-to-image alignment,
    such that for a suitable textual description $T$ of any concept $\Concept{C}\in\mathcal{C}$ one can reasonably assume $e(\Concept{C})\approx F_{\to L}(T)$.
    We use this to find embeddings: The embedding of a visual concept $\Concept{C}$ can be set to the DNN's text encoding $F_{\to L}(T)$ of a suitable textual description $T$ of $\Concept{C}$.
  \item \textbf{Existence of embeddings:}
    For all leaf concepts, embeddings $e(\Concept{C}_i)$ of sufficient quality exist in the latent space of $L$. 
  \item \textbf{Concentric distribution of subconcepts:}
    Representations of subconcepts are distributed in a \textbf{concentric manner} around its parent.
    Generally, this does not hold \cite{mikriukov2023gcpv},
    but so far turned out to be a viable simplification as long as semantic similarities are well preserved by the concept embedding function $e$ \cite{%
      ghorbani2019automatic,
      posada-moreno2024eclad
    }.
    I.e.\ for a superclass concept $\Concept{Parent}$ with children set $\mathcal{C}_S$ we can choose
    \begin{gather}
      \normalfont
      e(\Concept{Parent})\approx \operatorname*{mean}_{\Concept{Child}\in\mathcal{C}_S} e(\Concept{Child})
      \label{eq:superclass-is-mean}
    \end{gather}
  \item \textbf{Semantic interpolatability:}\label{ass:interpolatability}
    Consider a latent representation $v$ that is close to or inbetween (wrt.\ linear interpolation) some embeddings $e(C_i)$ and $e(C_j)$. We assume that $v$ can be interpreted to correspond to some concept, i.e.,
    $\exists \Concept{C}\in\mathcal{C}\colon \|e(\Concept{C})-v\|_2<\epsilon$ for some admissible error $\epsilon$.
    This is needed to make the averaging in the parent identification in \eqref{eq:superclass-is-mean} above meaningful.
  \end{enumerate}
\end{assumptions}
Note that Assumption~\ref{assumptions}\ref{ass:interpolatability} is very strong, stating that there is a correspondence between the semantic relations of natural language concepts, and the metric space structure of latent spaces. This is by no means guaranteed, but according to findings in word vector spaces \cite{%
  mikolov2013linguistic
} and also image model latent spaces \cite{fong2018net2vec} a viable assumption for the structure of learned semantics in DNNs.

\paragraph{Step 1: Obtain the embeddings $e(\Concept{C}_i)$.}
We here leverage the text-to-image alignment to directly define the concept-to-vector mapping $e$: 
$e(\Concept{C}) \coloneqq \operatorname{mean}_{x\in\Fun{toDNNInput}(\Concept{C})}F_{\to L}(x)$.
Following \cite{wan2020nbdt,yuksekgonul2022posthoc}, the $\Fun{toDNNInput}$ function can be a mapping from concept to a single textual description \cite{yuksekgonul2022posthoc} or to a set of representative images \cite{wan2020nbdt}.
\begin{itemize}[nosep]
\item \textit{Textual concepts:}
    The naive candidate for a textual description $\Fun{toDNNInput}(\Concept{C})\coloneqq \Fun{toText}(\Concept{C})$.
    However, some additional prompt engineering may be necessary, i.e., manual adjustment and finetuning of the formulation \cite{%
    ge2023improving,
    radford2021learning
    }.
    For example, following \cite{radford2021learning}
    we replace \enquote{\Concept{C}} by \enquote{an image of \Concept{C}} for the prompting.
\item \textit{Visual concepts:}
    Here we take the graphical $\Fun{toImages}(\Concept{C})$ specification of our concept.
    One could then employ standard supervised c-XAI techniques to find a common representing vector for the given images, e.g., as the weights of a linear classifier of the concept's presence \cite{%
    fong2018net2vec,
    kim2018interpretability
    }.
    We here instead simply feed the DNN with each of the images and capture its respective intermediate latent representations, which is valid due to the concentricity assumption.
\end{itemize}
If the text-to-image alignment is low, we found image representations of concepts to yield more meaningful results.

\paragraph{Step 2: Hierarchical clustering.}
Employ any standard hierarchical agglomerative clustering technique to find a hierarchy of partitionings of the set of given concept embeddings.
Each partitioning level represents one level of superclasses, with one cluster per class (see the simple example in \autoref{fig:overview}).
As of \eqref{eq:superclass-is-mean}, the mean of the cluster's embedding vectors is the embedding of its corresponding superclass (the \emph{cluster center}).

Note that the hierarchical clustering in principle allows to: (a) start off with more than one vector per leaf concept, e.g., coming from several image representations or from jointly using embeddings from textual and image representations; (b) weight the contribution of each child to the parent. This, however, is only viable together with means to automatically determine the weights, and not further pursued here.

\paragraph{Step 3: Decoding of cluster centers.}
We here use a two-step search approach to assign each cluster center a concept from the concept bank $\mathcal{C}$.
Given a cluster center $p$, the first optional step is to reduce the search space by selecting a subset of candidate concepts from $\mathcal{C}$. Following \cite{yuksekgonul2022posthoc},
(1a) we collect for every leaf concept $\Concept{C}$ the set of those concepts that, according to the ConceptNet knowledge graph \cite{speer2017conceptnet},
are related to $\Concept{C}$ by any of the relations in $\mathcal{R}_{\text{concepts}}=\{\Fun{hasA}, \Fun{isA}, \Fun{partOf}, \Fun{HasProperty}, \Fun{MadeOf}\}$:
\begin{gather}
  \Fun{Parents}(\Concept{C}) \coloneqq \{\Concept{P} \mid \bigvee_{\mathclap{R\in\mathcal{R}_{\text{concepts}}}} R(\Concept{P}, \Concept{C})\}
  \;.
\end{gather}
(1b) The union $\mathcal{P}=\bigcup_{\Concept{C}\text{ leaf in cluster}}\Fun{Parents}(\Concept{C})$ of these sets serves as candidate set for $p$.
Note that this is a simplification that allows to capture as superclass any best fitting commonality between the leaf concepts (e.g., background context like \Concept{indoor} or biological relation like \Concept{mammal} for $\{\Concept{cat},\Concept{dog}\}$ as in \autoref{fig:example-commitment}). Generally, there is a trade-off between very specific relation definitions, and fidelity to the learned knowledge of the model. The trade-off can be controlled by the broadening or narrowing of the candidate set. The here chosen broad definition of the \Concept{IsSuperClass} relationship between concepts favors fidelity to the model's learned knowledge. Investigating effects of more narrow concept candidate sets is future work.
(2) In the second step, the concept for $p$ is then selected from the candidate set $\mathcal{P}$ to be the one with minimum distance embedding (embeddings again obtained as in Step 1):
$e^{-1}(p)\coloneqq \argmin_{\Concept{P}\in\mathcal{P}}\|p-e(\Concept{P})\|_2$.

The final result then is a hierarchy tree, where leaf nodes are the originally provided concepts, inner nodes are the newly extracted superclasses, and the connections represent the \Concept{IsSuperclassOf} relation.
In the experimental section we will more closely investigate the influence of the proposed variants with/without prompt engineering and with/without finetuning.

\subsection{Inference of an ontology}\label{sec:approach-inference}

The such obtained ontology can be used for outlier-aware inference, i.e., classification of new input samples to one of the leaf concepts.
This will be useful not only as an interesting standalone application in safety-relevant classification scenarios,
but in particular for the validation.

The baseline of the inference is the $k$-nearest neighbor classifier: It directly compares the latent representation of a new input with each available concept embedding; and then assigns the majority vote of the $k$ nearest concept embeddings.
To enrich the inference process with information from the ontology, one instead traverses the ontology tree, at each node branching off towards the closest child node.
\begin{remark}
Note that this allows to easily insert an outlier criterion: If at a parent class $\Concept{P}$ none of the children nodes is closer than a threshold, the sample is considered an outlier \emph{of class \Concept{P}}. This neatly preserves the maximum amount of information available about the properties of the sample, and, thus, eases subsequent handling of the unknown input. For example, an outlier of (parent-)class \Concept{StaticObject} should be treated differently than one of (parent-)class \Concept{Animal}.
\end{remark} 

Hyperparameters of this inference procedure are the choice of similarity, including whether to take into account the size (variance/width) of the cluster, e.g., by favoring wide over near-to-point-estimate clusters; and the threshold for being an outlier.

\subsection{Validating and comparing learned ontologies}\label{sec:approach-validation}
We now get to the core goal of this paper: Verify or validate a given DNN using QR. For this we start with validation of an extracted ontology from \autoref{sec:approach-extraction}, and discuss how to measure its fidelity to DNN learned knowledge, and  alignedness to human prior knowledge, which here corresponds to the expected image-to-concept matching. Lastly, we show how one can encode a given ontology as contextualized embeddings to verify a DNN against given prior knowledge from QR.

\paragraph{Human-alignedness.}
One main desirable of a DNN's ontology is that it well aligns with the semantics that humans would expect and apply for the respective task. Any mismatch may either bring insights to the human on alternative solutions, or, more probably, indicates a suboptimal solution or even Clever Hans effect of the learned representations.
A straight-forward way to measure the human-alignedness is to test the \textbf{prediction accuracy} of the ontology when used for inference (see \autoref{sec:approach-inference}) on human-labeled samples. If human labels deviate often from the predictions, this indicates a bad alignment of the semantics the DNN has learned for the concepts from those a human would expect.
Other means to estimate the human-alignedness (not yet investigated in this work) are direct qualitative user studies, where human evaluators \textbf{manually check} the consistency of the obtained ontology tree with their own mental model; 
or \text{automatic checking} of consistency against given world knowledge or common sense ontologies like Cyc \cite{lenat1989building}.
Lastly, the improvement in humans' predictions about the behavior of the model, a typical human-grounded XAI metric \cite{schwalbe2023comprehensive},
could quantify in how far humans can make sense of the ontology.

A different aspect of human-alignedness is how well the ontology, in particular the inference scheme it defines, generalizes to novel concepts (semantic outliers) that so far have not occurred in leaves or nodes. The gerenalization can be measured as the performance in assigning a correct parent node.
A special case here are blended cases where the novel concept unifies features of very different classes, such as a \Concept{cat with wheel as walking support}. The uncertainty of the model in such blended cases can be qualitatively compared against human one, potentially uncovering a bias. 

\paragraph{Text-to-image alignment.}
The to-be-expected performance of cross-modal inference of the ontology (i.e., ontology defined using textual concepts, but inference done on images) directly depends on the quality of the text-to-image alignment. This motivates a use as an indicator for suboptimal text-to-image alignment.

\paragraph{Fidelity.}
Fidelity of the ontology, respectively shortcomings in the simplified modeling of the ontology, can be measured by the deviation between the baseline inference directly on the leaves, and the ontology inference.
Inference on the leaf concepts $\Concept{C}_i$ means we predict for an image $x$ the output class $\Concept{C}$ for which the textual embedding is closest to the embedding of $x$, proximity measured with respect to some distance $d$ (here: cosine similarity):
\begin{gather}
 \Concept{C} \coloneqq \argmin_{\Concept{C}\in (\Concept{C}_i)_i} d\left(
    F_{\to L}(\Fun{toText}(\Concept{C}), F_{\to L}(x))
 \right)
 \label{eq:naive-zero-shot}
\end{gather}
This is referred to as naive \emph{zero-shot} approach, following research on using foundation models on specialized tasks without finetuning (=with training on zero samples) \cite{%
ge2023improving,
radford2021learning
}.
%
The reason to choose this as a baseline is that the ideal tree should sort samples into the same leaf neighborhood as direct distance measurement would do. Simplifications that may infringe this equality are unequal covariances ($\approx$ widths) of sibling class clusters; the chosen similarity measure; or assuming perfect text-to-image alignment.

\paragraph{Verification against a given ontology.}
The previous extraction techniques yield an inspectable representation of the ontology learned by a model. This allows manual validation of the learned knowledge against models from QR.
Alternatively, one could directly verify a multimodal model against consistency with a given ontology: In short, we propose to modify the leaf concept embeddings from Step~1 such that they additionally encode their local part of the ontology, i.e., information about all desired parents of the leaf, as \emph{context}.
One can then measure the performance of naive inference (see \autoref{sec:approach-inference}) on these contextualized leaf nodes as defined in \eqref{eq:naive-zero-shot}.
A higher performance then means a better alignment of the context of a leaf concept with its image representations. This even would allow to narrow down unalignedness to specific concepts (those with bad inference results).
We suggest as point of attack for contextualization is the textual encoding: Let \Concept{C} be a leaf concept at depth $d$ in the tree with chain of parents $(\Concept{P}_i)_{i=1}^{d}$ from root to leaf. We can now follow \cite{ge2023improving} and modify the original $\Fun{tT}=\Fun{toText}$ function of a leaf concept to:
\begin{gather}
\Fun{toText}'(\Concept{C}) \coloneqq \text{\enquote{$\Fun{tT}(\Concept{P}_1), \dots, \Fun{tT}(\Concept{P}_d), \Fun{tT}(\Concept{C})$}}
\end{gather}
E.g., \Concept{cat} may turn into \textit{\enquote{\Concept{animal}, \Concept{pet}, \Concept{cat}}}.
The effect is that the obtained embedding (possibly after prompt finetuning as above) is shifted towards including the desired context; 
and all leaves together encode the complete ontology.

\section{Experiments}

\subsection{Settings}

\paragraph{Models under test.}
In our experiments, we utilized \textbf{CLIP} \cite{radford2021learning}, one of the first multimodal foundation model family accepting both text and images \cite{FoundationModelsReport2023}.
For text-to-image alignment CLIP was trained to map an image and its corresponding text descriptions onto a similar (with respect to cosine similarity) latent space representation. 
This general-purpose model captures rich semantic information, and achieves impressive performance compared to task-specific models across various applications, including image captioning \cite{%
barraco2022unreasonable,
cho2022finegrained%
}, recognition of novel unseen objects \cite{Amini-Naieni_2023_BMVC}, and retrieval tasks \cite{baldrati2022effective, sultan_2023}.
This makes it a common choice as basis for training or distilling more specialized models \cite{%
cho2022finegrained,
FoundationModelsReport2023
}, 
and thus a highly interesting target for validation and verification of its learned knowledge and internalized QR.
In our experiments, we explored various CLIP backbones, including ResNet-50, as well as Vision Transformer (ViT) variants featuring different patch sizes and model capacities (e.g., ViT-B/32, ViT-L/14)\footnote{Pre-trained models and weights were obtained from: \url{https://github.com/openai/CLIP}}.

\paragraph{Dataset.}
The \textbf{CIFAR-10} dataset \cite[Chap.\,3]{alex2009learning} is a benchmark in the field of computer vision, consisting of 60,000 32$\times$32 color images, split into 50,000 training and 10,000 test images. The images are equally distributed onto the 10 diverse classes \Concept{airplane}, \Concept{ship}, \Concept{car}, \Concept{truck}, \Concept{bird}, \Concept{cat}, \Concept{dog}, \Concept{deer}, \Concept{horse}, \Concept{frog}.
The choice of classes suits our initial study well, as they both exhibit pairs of semantically similar objects (e.g., \Concept{car}, \Concept{truck}), as well as mostly unrelated ones (e.g., \Concept{car}, \Concept{cat}), so we can expect a deep class hierarchy.
In our study, we conduct inference both of the baseline (naive zero-shot) and the proposed method on the CIFAR-10 test dataset \cite{alex2009learning}.

\paragraph{Fidelity baseline.}
As discussed in \autoref{sec:approach-validation}, the inference on the leaf concepts (\textbf{naive zero-shot} approach) serves as baseline (maximum performance) for fidelity measurements. The closer the tree inference gets to the naive zero-shot performance, the higher the fidelity.
We here choose as distance metric the cosine distance $\text{CosDist}(a,b) \coloneqq 1- \tfrac{a\cdot b}{\|a\|\cdot\|b\|}$ (0 for $a,b$ parallel, 1 for orthogonal, 2 for $a=-b$), going along with the training of CLIP.

\paragraph{Metrics.}
Any quantitative classification performances are measured in terms of \textbf{accuracy} of the results on CIFAR-10 test images against their respective ground truth label.

\begin{figure}[tb]
\centering
\hspace*{-.05\columnwidth}%
\includegraphics[width=1.05\columnwidth]{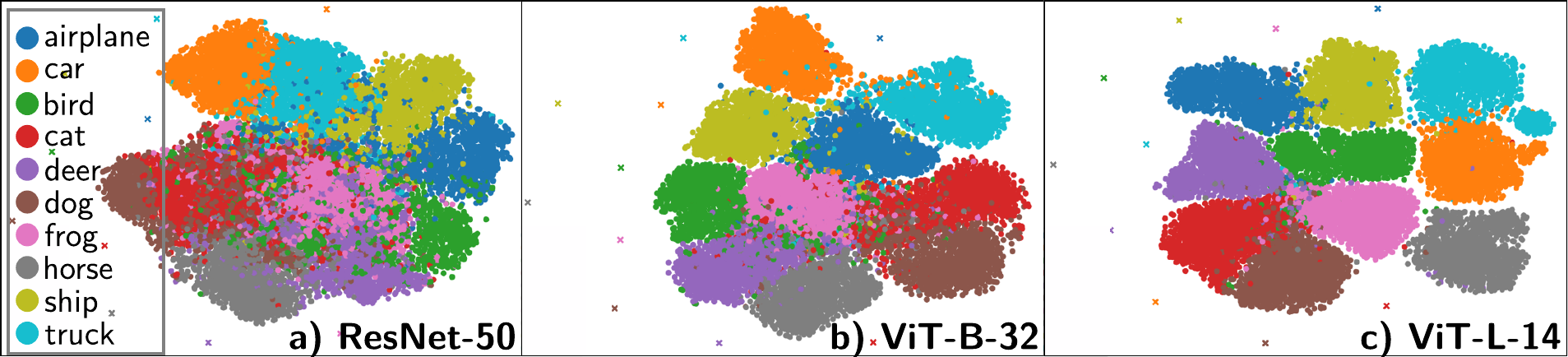}
\caption{Visualization of the latent space representations of CIFAR-10 embeddings in different CLIP model backbones (one color per class), generated using the distance-preserving t-SNE dimensionality reduction method \cite{maaten2008visualizing}.
The better class separation in the transformer-based backbones (\textbf{b)}, \textbf{c)}) are consistent with fidelity and human-alignedness results in Tabs.~%
\ref{tab:prompt-engineering}, 
\ref{tab:text-vs-image-encoding}.
}
\label{fig:LatentClip}
\vspace*{1.5\baselineskip}
\end{figure}

\subsection{Ablation Study: Influences on Human-Alignedness and Fidelity of Ontology Extraction}

As detailed in \autoref{sec:approach-validation}, to measure the \textbf{human-alignedness} of the given multi-modal encoder model, we evaluated the performance when using our extracted ontology for inference of class labels on new images. 
%
And as a \textbf{fidelity indicator}, we measure the performance drop between inference on the leaves (naive zero-shot approach) against that of inference on our tree.\footnote{%
Performance against a ground truth is only a proxy; future experiments should directly compare predictions of the two.}
Both are measured in the course of an ablation study to identify the influence of different settings on the ontology's usefulness and quality.


\paragraph{Investigated influences.}
Both the ontology extraction by means of agglomerative hierarchical clustering (see \autoref{sec:background-clustering}, as well as later the inference on new samples (see \autoref{sec:approach-inference}) rely on measuring similarities between embedding vectors.
However, due to being automatically optimized, the embeddings' optimal similarity metric is unknown.
Hence, we treat each choice of similarity metric as a hyperparameter, and investigate their influence on human-alignedness of the extracted ontology:
\begin{itemize}[nosep]
\item \textbf{Affinity:}
Affinity typically influences which data points are most similar, i.e., closest related, in the final tree structure.
In our experiments, we tested the standard Manhattan ($L_1$), Euclidean ($L_2$, and cosine distances.
\item \textbf{Linkage:}
This parameter determines the criterion used to merge clusters during the hierarchical clustering process, and in particular affects the shape and compactness of the clusters.
In our experiments, we tested the standard settings of Ward, complete, average, and single linkages.
Ward linkage minimizes the variance within clusters, while complete / average / single linkage focuses on the maximum / average / minimum distance between clusters.
\item \textbf{Inference similarity:}
We use use the same choices as for affinity.
\end{itemize}
Next, we compare different settings for obtaining the leaf embeddings. The following variants are considered: 
\begin{itemize}[nosep]
\item \textbf{Prompt tuning:}
In case text embeddings are to be obtained, CLIP suggests using text  prompts in the form \textit{\enquote{a photo of a \Concept{classname}}} rather than simply \textit{\enquote{\Concept{classname}}}, because the model is trained on image captions as text. If applied, this augmentation is done for both leaf and parent node textual embeddings.
\item \textbf{Text encoding vs. few-shot image encoding:}
As described in \autoref{sec:approach-extraction}, Step 1, the two different approaches to obtain leaf embeddings are text encoding and image encoding. We here only consider few-shot image encoding, i.e., specifying the concept via $<10$ images, which ensures manageable complexity of the hierarchical clustering algorithm\footnote{%
Standard implementations have a complexity of $\mathcal{O}(n^3)$ for $n$ leaf samples.}.
\end{itemize}



\paragraph{Results.}
An illustrative example of an ontology extracted from CLIP (ViT-L-14 backbone) using the prompt \textit{\enquote{a photo of a \Concept{classname}}} is provided in \autoref{fig:hierarchicalTree} for found-to-be-optimal settings according to the ablation study.
%
Consistently optimal hyperparameter settings with respect to human-alignedness and fidelity turned out to be affinity=Manhattan, linkage=complete, and inference similarity=cosine, which were also used to create the remainder of the ablation studies.
The accuracy results on CIFAR-10 of inference using the extracted ontology versus the naive-zero shot approach as a baseline for fidelity are given in Tabs.
\ref{tab:prompt-engineering} for the prompt engineering settings, and
\ref{tab:text-vs-image-encoding} for the comparison of text and image encodings of the leafs.
\\\emph{Please note that we did not yet conduct a cross-validation, 
so results should foremostly serve as guide for further investigations.}

\begin{table}[tbh]
\centering
\caption{Comparison of inference accuracy using naive zero-shot (Naive) and our method across different model architectures and textual prompt types. 
Fidelity calculated as ratio $\frac{\text{ours}}{\text{naive}}\in[0,1]$; best models marked.
}
\label{tab:prompt-engineering}
\vspace*{1.5\baselineskip}
\begin{tabular}{|l|cc>{\itshape}c|cc>{\itshape}c|}
\hline
\multicolumn{1}{|c|}{\textbf{}} & \multicolumn{6}{c|}{\textbf{Prompts}}                                                                                                                                                                                                                                                                                                      \\ \cline{2-7} 
                                & \multicolumn{3}{c|}{\textit{\enquote{\Concept{classname}}}}                                                                                                                            & \multicolumn{3}{c|}{\textit{\enquote{a photo of a \Concept{classname}}}}                                                                                      \\ \cline{2-7} 
\multicolumn{1}{|c|}{\textbf{}} & \multicolumn{1}{c|}{\textbf{\begin{tabular}[c]{@{}c@{}}Naive\end{tabular}}} & \multicolumn{1}{c|}{\textbf{\begin{tabular}[c]{@{}c@{}}Ours\end{tabular}}} & \multicolumn{1}{c|}{{\itshape~ratio~}} & \multicolumn{1}{c|}{\textbf{\begin{tabular}[c]{@{}c@{}}Naive\end{tabular}}} & \textbf{\begin{tabular}[c]{@{}c@{}}Ours\end{tabular}} & \multicolumn{1}{c|}{{\itshape~ratio~}}\\ \hline
\textbf{ResNet-50}              & 0.70                                                                                     & 0.46 &  0.66                                                         & 0.69                                                                                     & 0.67 &   0.97                                                        \\ \cline{1-1}
\textbf{ViT-B-32}               & 0.87                                                                                     & 0.82 &  \textbf{0.94}                                                         & 0.89                                                                                     & 0.85                                                          & \textbf{0.96} \\ \cline{1-1}
\textbf{ViT-L-14}               & \textbf{0.91}                                                                                     & \textbf{0.85} &  \textbf{0.93}                                                         & \textbf{0.95}                                                                                     & \textbf{0.92} &  \textbf{0.97}                                                          \\ \hline
\end{tabular}
\end{table}

\paragraph{First findings.}

In advance we manually validated the assumption of a good text-to-image alignment (Assumption~\ref{assumptions}\ref{ass:text-to-image-alignment}). For this we visualized the distribution and class separability of text and CIFAR-10 test sample embeddings in the latent spaces of the different CLIP backbones, results shown in \autoref{fig:LatentClip}.
The dimensionality-reduced visualizations suggest that with increasing parameter number, the clusters of different classes become more distinctly separated; and transformer-based backbones demonstrate superior separation. 
Notably, across all backbones, the text inputs and images are encoded in separate regions of the latent space, indicating a clear distinction between these two modalities in the model's internal representation.

The \result{prompt engineering}, i.e., replacing the text prompt \enquote{\Concept{classname}} with \enquote{\textit{a photo of \Concept{classname}}} turned out to be have a strong \result{positive impact on human-alignedness and fidelity} in case of the worse aligned CNN-based CLIP backbone, and still a notable one for the already good transformer backbones.

In contrast, using \result{few images instead of text to obtain the leaf embedding resulted in worse performance}. However, in our initial tests performance seemed to increase with the number of images: Dropping the few-shot constraint showed competitive results. In the following table, we replaced the leaf node information with the randomly-sampled training images in the respective class. 

\begin{table}[tbh]
\centering
\caption{
Comparison of inference accuracy for different ways to obtain the leaf embeddings: \emph{few-shot} image embeddings vs. textual embeddings (\emph{zero-shot}), with the naive zero-shot approach as baseline. Best model \textbf{bold}.
}
\vspace*{1.5\baselineskip}
\label{tab:text-vs-image-encoding}
\begin{tabular}{|l|ccc|cl|}
\hline
                                & \multicolumn{3}{c|}{\textbf{Few-Shot}}                                                         & \multicolumn{2}{c|}{\textbf{Zero-Shot}}                                  \\ \cline{2-6} 
\multicolumn{1}{|c|}{\textbf{}} & \multicolumn{1}{c|}{\textbf{1-shot}} & \multicolumn{1}{c|}{\textbf{5-shot}} & \textbf{10-shot} & \multicolumn{1}{c|}{\textbf{Naive}} & \multicolumn{1}{l|}{\textbf{Ours}} \\ \hline
\textbf{ResNet-50}              & 0.45                                 & 0.58                                 & 0.61             & 0.69                                & 0.67                               \\ \cline{1-1}
\textbf{ViT-B-32}               & \textbf{0.67}                                 & \textbf{0.79}                                 & \textbf{0.86}             & 0.89                                & 0.85                               \\ \cline{1-1}
\textbf{ViT-L-14}               & 0.64                                 & 0.76                                 & 0.80             & \textbf{0.95}                                & \textbf{0.92}                               \\ \hline
\end{tabular}
\end{table}

It should be noted, that a better performance of the textual embedding could possibly be attributed to a sub-optimal text-to-image alignment. This would be consistent with the insights into the distribution and class separability of image and text embeddings in the latent space in \autoref{fig:LatentClip} (with respect to Euclidean distance).
It should be further investigated, whether this must be attributed to disparity in metrics, the domain shift to CIFAR-10 inputs, or could serve as an indicator for bad text-to-image alignment wrt.\ the considered classes.

\subsection{Ontology validation and verification}

\paragraph{Validation: qualitative results.}
A manual inspection of the obtained ontologies (see \autoref{fig:hierarchicalTree} for an example) showed, that \result{good human-alignedness also coincides with seemingly valid tree structures}.
Seemingly valid here means, that a human inspector can easily find convincing arguments for the validity most of the splitting criteria of the nodes.
In \autoref{fig:hierarchicalTree},
two trees which are created with different parameters are compared. The tree on the left, which uses ViT-L/14 as a backbone, affinity clustering, and Manhattan linkage, achieves 92\% accuracy on the classification task. In contrast, the tree on the right, created with a ResNet-50 backbone, affinity clustering, and Euclidean linkage, yields an accuracy of 45\%. One of the reasons for the low accuracy score in the classification task for the tree on the right is that its decision process does not align well with human-like decision-making. For example, the structure first checks whether an object is a \textit{"vehicle"} and then whether it is \textit{"meat"}. This decision process deviates from human-aligned reasoning, which can also be observed through manual inspection.
%

Furthermore, we identified the tendency that the \result{superior vision transformer backbones also showed the seemingly more valid tree structures}.
This \result{possible architectural dependency of good ontological commitment} should be further investigated.

\paragraph{Verification against a given ontology.}
To exemplify the verification of ontological commitment against a given ontology, we chose the simple tree structure provided by \cite{wan2020nbdt} for CIFAR-10 dataset. To label the inner nodes of this tree, we utilized two external knowledge sources: WordNet \cite{fellbaum2010wordnet} and GPT-4 \cite{achiam2023gpt}, in each case bottom-to-top queried for a textual description of a parent for sibling nodes.
We then used the ontology information to create contextualized leaf embeddings, as described in \autoref{sec:approach-validation}, and applied naive zero-shot inference on these contextualized leaves.
For WordNet, we labeled each node with the closest matching superclass. For GPT-4, we queried the model to provide the superclass of the given leaf nodes.

Initial verification results for the different given ontologies are shown in \autoref{tab:KnowdgeInsertion}:
As expected, using the extracted learned ontology for the contextualization caused no change compared to the baseline of non-contextualized embeddings; this contextualization is supposed to be equivalent to the non-contextualized leaf embeddings from the perspective of the model.
However, the contextualization with external ontologies caused a strong drop in inference accuracy.
A closer look at the results showed that those leaves with parents mentioning technical terms (e.g., \enquote{non-mammalian vertebrate}) were mostly misclassified, indicating that the learned knowledge is inconsistent / not aware of these parts of the given ontologies. 
Further research is needed on practical implications (e.g., thus induced error cases), and how to align the ontologies.

\begin{table}[tbh]
\centering
\caption{
Verification results of different models against different sources of external ontologies:
the NBDT tree structure~\cite{wan2020nbdt} with \emph{WordNet}~\cite{fellbaum2010wordnet} or \emph{GPT-4}~\cite{achiam2023gpt} queried node labels; 
versus no contextualization (\emph{Naive}) and contextualization against the extracted ontology (\emph{Ours}).
Values are measured in inference accuracy on contextualized nodes.
}
\label{tab:KnowdgeInsertion}
\vspace*{1.5\baselineskip}
\begin{tabular}{l|cc|cc|}
\cline{2-5}
                                        & \multicolumn{1}{c|}{\textbf{WordNet}} & \multicolumn{1}{c|}{\textbf{GPT-4}} & \multicolumn{1}{c|}{\textbf{Naive}} & \multicolumn{1}{c|}{\textbf{Ours}}             \\ \hline
\multicolumn{1}{|l|}{\textbf{ResNet-50}} & 0.31             &  0.36          &  0.69          & 0.67 \\ \hline
\multicolumn{1}{|l|}{\textbf{ViT-B-32}} & 0.40             & 0.53           & 0.89           & 0.85 \\ \hline\multicolumn{1}{|l|}{\textbf{ViT-L-14}} & 0.52             &     0.54        &  0.95          & 0.92  \\ \hline
\end{tabular}
\end{table}


\section{Future work: Applications and next steps}

\subsection{Applications of learned ontology extraction}
Our method opens up several further interesting applications for the use of QR in DNN understanding, verification, and improvement.
\paragraph{Optimal learned reasoning representations.}
As discussed above, access to the internal ontology of a DNN is key to understand its internal QR. In particular, an open research question is,
\emph{what kind of concept representations are DNNs optimized for}, and, subsequently, 
\emph{which kinds of reasoning would be supported by this?} For example, qualitative spatial reasoning 
would most benefit from a region-based representation of concepts,
while cone-based reasoning from cones as representations \cite{ozcep2023embedding}.
The quantitative measurement of ontological commitment allows to do ablation studies on different representations of concepts and relations, e.g., different similarity measures.

\paragraph{DNN inspection.}
The obtained ontologies open up new inspection possibilities for DNNs.
An interesting one could be to generate \textbf{contrastive examples} \cite{guidotti2022counterfactual}: 
Change a given input minimally such that the class/superclass changes, possibly under a constraint to remain within a given superclass.
Also, one could globally test the models against biases towards scenerios respectively background.
A bias is uncovered, if the commonality of two classes is based on background rather than functionally relevant features;
possibly supported on test samples generated by inpainting techniques. 
Unfortunately, the text-to-image alignment training of foundation models may easily introduce such a bias, as concepts occurring in similar image scenarios additionally will occur in similar textual context.
E.g., one may expect \Concept{cat} and \Concept{dog} to be similar, as both often occur indoors.

\paragraph{Knowledge insertion.}
The final goal of the introspection discussed above should be to not only be able to verify the learned ontological commitment, but also to control both the commitment, and subsequently the learned reasoning.
This might be achieved by adding penalties during training, determined by iterative ontology extraction and model finetuning.
Thus, a foundation model with acceptable ontological commitment may be obtained.
Lastly, to distill this knowledge of the large model into smaller specialized models,
standard model distillation techniques could be amended \cite{papernot2016distillation}.
Concretely, regularization terms can be added to (1) enforce that correspondences to some/most of the concepts, and to (2) enforce respective similarities and other relationships between the concepts.

\subsection{Next steps} 
Our initial experiments are clearly limited in their extend, so immediate next steps should encompass more experiments on measuring \textbf{human-alignedness} respectively a larger ablation study on possible influence of the made assumptions.
Such can be domain shifts, like text-to-image, and real-to-synthetic image.
Experiments should include user studies, and comparison to existing ontologies;
Similarly, the \textbf{outlier detection and handling} capabilities of ontologies should be further investigated, both for novel as well as novel blended classes.
Lastly, 
it can be investigated how to extend the here proposed approach \textbf{from multimodal models to unimodal} ones, allowing to compare the ontologies of large foundation models against that of state-of-practice small and efficient object detectors.

\section{Conclusion} 
Altogether, this paper tackles the problem how to validate and verify a multimodal DNN's learned knowledge using QR. Concretely, we take the step to unveil the ontological commitment of DNNs, i.e., the learned concepts and (here: superclass-)relations.
For this, we proposed a simple yet effective approach to (1) uncover yet undiscovered superclasses of given subclasses as used by the DNN; and to (2) extract a full hierarchical class tree with the \Concept{IsSuperClass}-relationships; together with means to verify and validate the extracted part of the learned ontology.
Even though this initial proof-of-concept still relies on some simplifications, our initial experiments could already extract meaningful class hierarchies from concurrent multimodal DNNs, and reveal inconsistencies with existing ontologies. 
These may serve as a basis to access further insights into the ontological commitment of DNNs, and subsequently validate and verify its learned QR. We are confident that, eventually, this could allow to control, i.e., correct and integrate, valuable prior knowledge from QR into DNNs, creating powerful yet verifiable and efficient hybrid systems. Thus, we hope to spark further interest into interdisciplinary research of QR for verification of DNNs within the QR community.



\begin{ack}
The paper was written in the context of the "NXT GEN AI Methods" research project funded by  
the German Federal Ministry for Economic Affairs and Climate Action (BMWK), The authors would like to thank the consortium for the successful cooperation.
\end{ack}



\bibliography{mybibfile}

\end{document}